\documentclass[conference]{IEEEtran}
\IEEEoverridecommandlockouts
\usepackage{cite}
\usepackage{amsmath,amssymb,amsfonts}
\usepackage{algorithmic}
\usepackage{graphicx}
\usepackage{textcomp}
\usepackage{xcolor}
\usepackage{array}
\usepackage{booktabs}
\usepackage{float}
\usepackage{multirow}

\def\BibTeX{{\rm B\kern-.05em{\sc i\kern-.025em b}\kern-.08em
    T\kern-.1667em\lower.7ex\hbox{E}\kern-.125emX}}
\begin{document}

\title{Towards Resilient and Efficient LLMs: \\ A Comparative Study of Efficiency, Performance, \\ and Adversarial Robustness \\
}

\author{\IEEEauthorblockN{Xiaojing Fan*}
\IEEEauthorblockA{\textit{New York University}\\
xf435@nyu.edu}
*Corresponding Author
~\\
\and
\IEEEauthorblockN{Chunliang Tao}
\IEEEauthorblockA{\textit{New York University}\\
ct1942@nyu.edu}
}
\IEEEpeerreviewmaketitle
\maketitle

\begin{abstract}
With the increasing demand for practical applications of Large Language Models (LLMs), many attention-efficient models have been developed to balance performance and computational cost. However, the adversarial robustness of these models remains under-explored. In this work, we design a framework to investigate the trade-off between efficiency, performance, and adversarial robustness of LLMs and conduct extensive experiments on three prominent models with varying levels of complexity and efficiency -- Transformer++, Gated Linear Attention (GLA) Transformer, and MatMul-Free LM -- utilizing the GLUE and AdvGLUE datasets. The AdvGLUE dataset extends the GLUE dataset with adversarial samples designed to challenge model robustness. Our results show that while the GLA Transformer and MatMul-Free LM achieve slightly lower accuracy on GLUE tasks, they demonstrate higher efficiency and either superior or comparative robustness on AdvGLUE tasks compared to Transformer++ across different attack levels. These findings highlight the potential of simplified architectures to achieve a compelling balance between efficiency, performance, and adversarial robustness, offering valuable insights for applications where resource constraints and resilience to adversarial attacks are critical.
\end{abstract}


\section{Introduction}
In recent years, large language models (LLMs) have grown rapidly and demonstrated unprecedented performance across various language understanding and generation tasks. These achievements are primarily driven by developments of transformer-based architecture \cite{vaswani2017attention, devlin2019bert, brown2020language} as well as innovations in training techniques and model scaling \cite{johnson2020efficient, rosset2020turing, huang2021gshard}. However, as LLMs grow in complexity and size, concerns about their efficiency and scalability have become increasingly prominent. The substantial computational resources required for training and deploying these models often limit their accessibility and practicality, especially in resource-constrained environments. Additionally, LLMs are shown to be vulnerable to adversarial attacks, raising critical concerns about the reliability and robustness of LLMs. Adversarial attacks involve imperceptible perturbations to the input data, leading to output inconsistencies and performance degradation \cite{zhang2020adversarial}.

The growing demand for practical and efficient applications of LLMs has spurred the development of models designed to balance performance with computational cost \cite{wan2024efficientlargelanguagemodels}. Some notable attention-efficient models, such as RetNet \cite{sun2023retentivenetworksuccessortransformer}, Mamba \cite{gu2024mambalineartimesequencemodeling}, Gated Linear Attention (GLA) Transformer \cite{yang2024gatedlinearattentiontransformers}, and MatMul-Free LM \cite{zhu2024scalablematmulfreelanguagemodeling}, have been developed by reforming sequence handling mechanism and are shown to reduce the computational burden without significantly compromising performance \cite{katharopoulos2020transformers, choromanski2020rethinking}. However, the adversarial robustness of these attention-efficient models remains unstudied, which is essential for extending their applications in sectors where reliability is crucial, such as healthcare and software security.

To the best of our knowledge, no comprehensive evaluation framework exists to assess the adversarial robustness of attention-efficient models. Previous studies have predominantly focused on performance and robustness of traditional transformer architectures, with little consideration of lightweight models designed for efficiency \cite{schwinn2023adversarial, kumar2023certifying, xhonneux2024efficient}. In this paper, we aim to fill the research gap by presenting a framework to assess trade-offs between computational efficiency, performance, and adversarial robustness of LLMs and applying our framework to three prominent models with different complexity -- Transformer++ \cite{thapak2020transformer}, GLA Transformer \cite{yang2024gatedlinearattentiontransformers}, and MatMul-Free LM \cite{zhu2024scalablematmulfreelanguagemodeling}. Transformer++ is an advanced architecture known for its high performance, while GLA Transformer and MatMul-Free LM are recent attention-efficient models with innovative attention mechanisms and proven computational efficiency. In our framework, we perform task-specific fine-tuning and evaluation of the three models on four NLP classification tasks from the GLUE dataset \cite{wang-etal-2018-glue}. Then we evaluate the fine-tuned models on the corresponding tasks from the AdvGLUE dataset \cite{wang2022adversarialgluemultitaskbenchmark}, which includes word-level, sentence-level, and human-level adversarial attacks to the GLUE dataset. Our experiments show that GLA Transfomer and MatMul Free LM achieve higher efficiency and comparative performances compared to Transformer++ across GLUE tasks. In addition, GLA Transfomer demonstrates superior robustness across all attack levels, while MatMul-Free LM is more robust to word-level attacks and equally robust to sentence-level and human-level attacks as Transformer++. These findings offer valuable insights into the applicability and resilience of these models, thereby informing future development and deployment strategies in different environments.

Our studies make several contributions to the field of LLMs:
\begin{enumerate}
    \item Our studies bridge the research gap on the adversarial robustness of attention-efficient models, such as GLA Transfomer and MatMul-Free LM, by assessing their resilience under different types of adversarial attacks.
    \item We introduce a novel framework and conduct the first empirical studies to assess trade-offs between computational efficiency, performance, and adversarial robustness of LLMs with varying complexity. 
    \item Our research informs strategies for and guides future studies in selecting and deploying LLMs in resource-constrained and adversarial environments.
\end{enumerate}

\section{Related Work}
\subsection{Vulnerability of LLM}
Large Language Models (LLMs) have demonstrated remarkable performances across a variety of tasks, driving revolutionary changes and developments in various domains such as art  \cite{deng2024composerx}, healthcare \cite{yang2023text, liu2023chatgptpoweredconversationaldrugediting, ding2024largelanguagemultimodalmodels}, software security \cite{zhang2024llamafuzzlargelanguagemodel, rasheed2024aipoweredcodereviewllms}, etc., some of where reliability and robustness are critical. Failures in such domains can lead to severe consequences, raising significant concerns about the reliability of LLMs. Despite their advanced capabilities, LLMs and related applications have been shown to possess vulnerabilities. For instance, Wallace et al. unveiled that subtle yet sophisticated modifications from human-generated adversarial examples can significantly mislead models to generate incorrect responses \cite{wallace2019trickcanhumanintheloopgeneration}. Retrieval-based Augmented Generative Models (RAG), which enhance LLMs by integrating external knowledge, can be manipulable if malicious content is ingested into their knowledge bases \cite{tan2024gluepizzaeatrocks}. To mitigate the vulnerabilities of LLMs, ongoing studies have been investigating various solutions, including but not limited to incorporating adversarial training techniques \cite{shafahi2019adversarialtrainingfree}, adopting meta-learning techniques \cite{tao2023metalearning} to adapt defenses against adversarial attacks, systematically rectifying LLMs via dead-end analysis\cite{cao2023systematic}, and integrating uncertainty-aware models to handle ambiguous inputs \cite{lu2024uncertainty}. However, existing approaches still fail to provide comprehensive protection against evolving forms of vulnerabilities, highlighting the need for ongoing research to address a wide range of threats and ensure the robustness of LLMs in diverse and dynamic environments.

\subsection{Adversarial Attacks}
Amid the proliferation of Artificial Intelligence (AI), various forms of adversarial attacks have been developed, challenging the robustness and questioning the security of AI. For example, DeepWordBug \cite{gao2018blackboxgenerationadversarialtext} demonstrates the vulnerability of deep learning classifiers by creating effective adversarial text through small character-level modifications such as insertions, deletions, and swaps, underscoring how easily these systems can be deceived. Similarly, covert injection attacks, which introduce malicious behavior into an otherwise benign language model to produce harmful outputs, emphasize the critical need for robust integrity checks on model responses \cite{tan2024wolfwithincovertinjection}. Furthermore, prompt injection methods can extract memorized training data by crafting input prompts that increase the likelihood of the model generating memorized content, highlighting significant privacy breach potentials. A recent sequential query-based black-box method also exploits model responses to generate effective adversarial examples, demonstrating attack effectiveness comparable to white-box approaches \cite{taoSQBA}. To facilitate the evaluate the robustness of LLMs against such diverse adversarial attacks, benchmark datasets like AdvGLUE are developed. AdvGLUE provides a comprehensive suite for testing models under various adversarial scenarios, aiding in the identification of vulnerabilities and the enhancement of the resilience of LLMs.

\subsection{Efficient LLMs}
The growing demand for large language models (LLMs) across various applications has spurred the development of more efficient models that balance performance with computational and memory efficiency. For example, DistilBERT \cite{sanh2020distilbertdistilledversionbert} and Tiny BERT\cite{jiao2020tinybertdistillingbertnatural} utilize model distillations to reduce the sizes of LLMs by training a ``student" model to replicate their ``teacher" model while retaining most performance. Q8BERT \cite{Zafrir_2019} employs quantization techniques to convert 32-bit floating numbers to 8-bit precision to make models lighter without significantly compromising performance. The Gated Recurrent Units (GRU) method \cite{cho2014propertiesneuralmachinetranslation} is introduced to address limitations in recurrent neural networks (RNNs) and demonstrates a promising balance between performance and efficiency in handling NLP tasks involving long sequences \cite{xu2024textsentimentanalysisclassification}. Innovative and scalable training strategies, such as MiniCPM \cite{hu2024minicpmunveilingpotentialsmall}, have also been introduced to maximize the potential of small LLMs while ensuring resource efficiency. In addition, alterations in attention mechanism have been applied to models including GLA Transformer \cite{yang2024gatedlinearattentiontransformers} and Matmul-Free LM \cite{zhu2024scalablematmulfreelanguagemodeling} to further reduce computational memory in all stages of model training and inference. GLA Transformer adopts the linear attention mechanism and a gating mechanism to reduce computational complexity and optimize memory usage, achieving a balance between complexity and efficiency. MatMul-Free LM fundamentally changes the computation paradigm by replacing matrix multiplication operations in the attention mechanism with ternary weights and element-wise operations, largely reducing computational complexity. Due to efficiency-oriented advancements, these models demonstrate significant potential for deployment in resource-constrained environments. Previous studies have primarily focused on assessing trade-offs between performances and efficiencies of these models, leaving their adversarial robustness under-explored. However, as the needs and applications of these efficiency-focused models grow, it is crucial to comprehensively assess their resilience to adversarial attacks and understand the trade-offs between efficiency, performance, and adversarial robustness.

\begin{figure*}[ht]
  \centering
  \includegraphics[width=\linewidth]{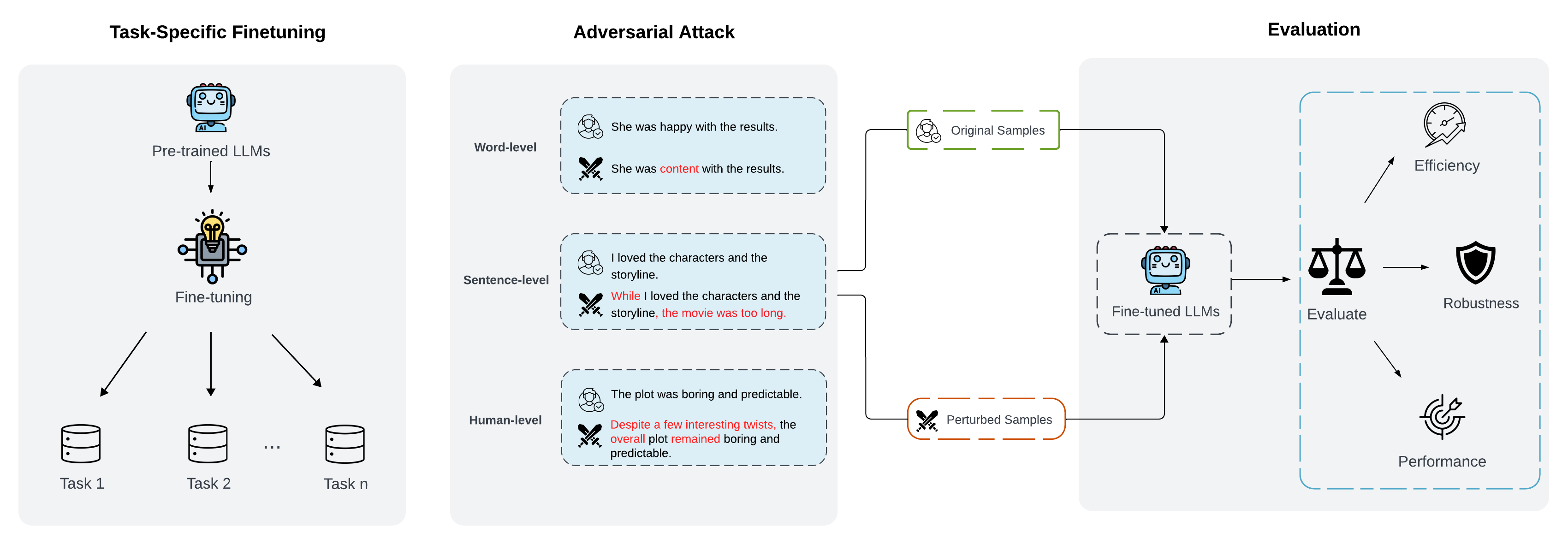}
  \caption{E-P-R Trade-off Evaluation Framework}
  \label{fig:evaluation_framework}
\end{figure*}

\section{Methodology}
\subsection{E-P-R Trade-off Evaluation Framework}
In this work, we propose the Efficiency-Performance-Robustness (E-P-R) Trade-off Evaluation Framework, as illustrated in Figure \ref{fig:evaluation_framework}, to comprehensively assess Large Language Models (LLMs) based on three critical aspects: computational efficiency, task performance, and adversarial robustness. This framework provides a structured approach to understanding the trade-offs between these aspects, facilitating informed decision-making when deploying LLMs in real-world applications. The framework consists of three key components: task-specific fine-tuning, adversarial attack generation, and trade-off evaluation.

\textbf{Task-specific Fine-tuning } The first part is fine-tuning pre-trained LLMs on a variety of language tasks to adapt the models for specific applications. This step ensures that each model is optimized for the specific task while maintaining comparability across models. The fine-tuned models are prepared for evaluation under both standard and adversarial conditions. 

\textbf{Adversarial Attack } To assess adversarial robustness, we introduce perturbations to the input data at various levels, including word-level, sentence-level, and human-level. These adversarial examples are generated by applying different types of perturbations to the original input samples from each task. The robustness of each LLM is then evaluated based on its performance on these adversarial examples, allowing for a detailed examination of how well the models handle malicious or unexpected input variations. 

\textbf{Trade-off Evaluation } Once the models have been fine-tuned and tested under adversarial conditions, we conduct a comprehensive analysis of the trade-offs between efficiency, performance, and robustness. Efficiency is measured in terms of computational resources required for fine-tuning and inference, while performance and robustness are evaluated on both standard and adversarial data. The adversarial robustness of each model is assessed based on its ability to maintain performance under various levels of adversarial attacks. By examining the interplay between these three aspects, our framework enables a nuanced understanding of the strengths and limitations of each model, providing insights for optimizing model deployment in different environments. 

\subsection{Large Language Models}
In this work, we focus on three language models: Transformer++ \cite{thapak2020transformer}, GLA Transformer \cite{yang2024gatedlinearattentiontransformers}, and MatMul Free LM \cite{zhu2024scalablematmulfreelanguagemodeling}. These models are carefully selected to represent distinct approaches to the core attention mechanism, which is central to transformer architectures and directly impacts model scalability and practicality across different application contexts. They have all demonstrated comparative performances on various language tasks with different training efficiency, but their adversarial robustness remains unexplored. Comparing these three models enables a holistic evaluation of the trade-offs between computational efficiency, performance, and adversarial robustness.

\textbf{Transformer++ } Transformer++ is an advanced architecture that builds upon the original transformer architecture \cite{vaswani2017attention}. Transformer++ follows the encoder-decoder architectures of Transformer while introducing a new form of multi-head attention that learns in-context dependencies through convolution. Specifically, Transformer++ employs convolution-based attention by using an adaptive sequence module and adopts an adaptive query module and a dynamic convolution module to capture local word context and entire sequence context, respectively. Given the total number of multi-head (\textit{H}) and weight matrices $W_i^Q, W_i^K$, and $W_i^V$, Equation (\ref{eq:mutilhead_attention}) shows how the model combines convolution-based attention with traditional self-attention that captures global dependencies to form a more comprehensive attention mechanism \cite{katharopoulos2020transformers}. Transformer++ has been demonstrated to be a strong transformer baseline model for various language tasks in previous studies.

\begin{equation}
\begin{aligned}
\text{MultiHead}(S) &= \text{Concat}(\text{head}_1, \ldots, \text{head}_H)W_O \\
\text{head}_{i, 1:H/2} &= \text{SelfAttention}(QW_i^Q, KW_i^K, VW_i^V) \\
\text{head}_{H/2:H} &= \text{DynamicConv}(S, t, d)
\end{aligned}
\label{eq:mutilhead_attention}
\end{equation}

\textbf{GLA Transformer } GLA Transformer builds upon the linear attention mechanism \cite{katharopoulos2020transformers}, which aims to reduce the computational complexity from which traditional Transformers suffer. Linear attention as shown in Equation (\ref{eq:linear_attention}) reduces computational complexity from \(O(n^2)\) to \(O(n)\) where $d_k$ represents the dimensionality of the keys, compared with traditional softmax attention as shown in Equation (\ref{eq:softmax_attention}). In addition, GLA Transformer integrates a data-dependent gating mechanism that modulates the attention scores dynamically based on input sequence to make the model selectively attend to important features and ignore less relevant ones, improving the efficiency and effectiveness of the attention process. GLA Transformer also implements a parallel chunk-wise form to handle large sequences by dividing them into non-overlapping chunks that can be processed in parallel and then combined. GLA Transformer has exhibited higher training speed and comparable performance to Transformer++ on various language tasks \cite{yang2024gatedlinearattentiontransformers}.

\begin{equation}
\text{LinearAttention}(Q, K, V) = \phi(Q) (\phi(K)^T \phi(V))
\label{eq:linear_attention}
\end{equation}

\begin{equation}
\text{Attention}(Q, K, V) = \text{softmax}\left(\frac{QK^T}{\sqrt{d_k}}\right)V
\label{eq:softmax_attention}
\end{equation}

\textbf{MatMul-Free LM} Traditional large language models (LLMs) rely heavily on matrix multiplication operations, which become a bottleneck as model size or data size increases. MatMul-Free LM addresses this issue by replacing matrix multiplications with more computationally efficient alternatives, such as simpler operations including additions and subtractions \cite{zhu2024scalablematmulfreelanguagemodeling}. To further optimize performance, ternary weights are incorporated into BitLinear modules \cite{wang2023bitnetscaling1bittransformers}, replacing traditional dense layers and significantly reducing memory and computation requirements. MatMul-Free LM also modifies the self-attention mechanism by using a variant of the Gated Recurrent Unit (GRU) to maintain the transformer architecture while reducing computational overhead. Additionally, MatMul-Free LM leverages the efficiency of hardware accelerators such as GPUs and FPGAs. By optimizing the computation steps and reducing the number of required operations, the model achieves substantial improvements in hardware utilization. These innovations collectively lead to substantial reductions in memory usage and faster training and inference times of MatMul-Free LM while maintaining comparable performance compared to traditional Transformer architecture \cite{zhu2024scalablematmulfreelanguagemodeling}.

\subsection{Dataset}
In this study, we utilize both the General Language Understanding Evaluation (GLUE) \cite{wang-etal-2018-glue} and Adversarial GLUE (AdvGLUE) \cite{wang2022adversarialgluemultitaskbenchmark} benchmark datasets to comprehensively evaluate the performance and adversarial robustness of large language models (LLMs). These datasets are specifically chosen for their ability to test a wide range of natural language processing (NLP) skills and their relevance to real-world language understanding tasks.  

GLUE is a widely recognized benchmark designed to evaluate the generalization capabilities of language models across a diverse set of NLP tasks. These tasks include sentiment analysis, paraphrase detection, and sentence similarity, among others, which collectively cover multiple facets of language understanding such as syntax, semantics, and inference. The primary reason for selecting GLUE is its well-established use as a standard benchmark for evaluating NLP models, which allows comparison with a broad set of results from the research community. GLUE's inclusion ensures that our evaluation covers a comprehensive range of language tasks, ensuring that our results reflect the overall language processing capabilities of the models under study. 

On the other hand, AdvGLUE extends the GLUE benchmark by introducing a set of 14 distinct adversarial attack methods that rigorously test model robustness against adversarial inputs. The rationale for incorporating AdvGLUE is to move beyond performance evaluation under standard conditions and assess how well the models perform when faced with adversarial challenges. Robustness to adversarial inputs is a critical consideration for real-world applications, especially in domains where language models might be deployed in environments prone to manipulation or adversarial attacks, such as automated customer service, security systems, or sensitive decision-making processes. AdvGLUE uses the same evaluation metrics as GLUE, allowing for a direct and fair comparison between performance on benign examples and adversarially perturbed inputs. This ensures a balanced view of both standard and adversarial model performance, a key focus of our research. 

In our study, we focus on four key tasks from both GLUE and AdvGLUE: Standard Sentiment Analysis (SST-2), Multi-Genre Natural Language Inference (MNLI), Quora Question Pairs (QQP), and Question Natural Language Inference (QNLI). These tasks are chosen not only because of their relevance to various real-world applications but also because they cover different aspects of language understanding and inference. By evaluating models on both standard and adversarial versions of these tasks, we are able to thoroughly assess their performance and robustness.

\section{Experiments}
\subsection{Experiment Setup}
We conduct experiments on the pre-trained Transformer++, GLA Transformer, and MatMul-Free LM to evaluate trade-offs between efficiency, performance, and adversarial robustness.

\textbf{Training and Fine-tuning } For a fair comparison, all three models are pre-trained on the SlimPajama dataset \cite{soboleva2023slimpajama} with 100B tokens and 1.3B model parameters. The pre-trained models are then fine-tuned on four classification tasks from the GLUE training set: SST-2, MNLI, QQP, and QNLI. Specifically, we fine-tune for 1 epoch on A100 GPU with a batch size of 8 and early stopping criteria for all models across all tasks. We use the AdamW optimizer \cite{loshchilov2019decoupledweightdecayregularization} and a cosine learning rate scheduler with an initial learning rate of 1e-5. Weight decay is set to 0.01 and gradient clip is set to 1.0. We observe that all models converge on all tasks before completing 1 epoch.

\textbf{Adversarial Attacks } We utilize the AdvGLUE dataset to perform adversarial attacks on the fine-tuned models. In the AdvGLUE tasks, word-level and sentence-level perturbations are applied to sample cases of corresponding tasks in the GLUE development set, and human-level perturbed samples are added as new samples.

\textbf{Model Evaluation } We measure the efficiency of each model in the fine-tuning stage and evaluate the performance of fine-tuned models on original samples from the GLUE development set and adversarial samples from the AdvGLUE dataset. Adversarial robustness of fine-tuned models is assessed by comparing performance differences between GLUE tasks and AdvGLUE tasks. Specific evaluation metrics are documented in the following section.

\subsection{Evaluation Metrics}
\textbf{Efficiency}
We quantify the fine-tuning efficiency of each model with the average GPU memory usage (\%).

\textbf{Performance}
We assess model performance with accuracy on GLUE tasks. Accuracy is given by the percentage of examples correctly classified by the model where $\hat{y}_i$ is the predicted label and $y_i$ is the true label. 

\begin{equation}
Acc = \frac{\sum_{i=1}^{N} \mathbf{1}(\hat{y}_i = y_i)}{N_{\text{total}}} \times 100\%
\end{equation}

\textbf{Adversarial Robustness}
To quantify the robustness of models against adversarial attacks, we introduce accuracy degradation percentage (ADP), given as 

\begin{equation}
    \begin{aligned}
\text{ADP} &= \left( \frac{\text{Acc}_{\text{GLUE}} - \text{Acc}_{\text{AdvGLUE}}}{\text{Acc}_{\text{GLUE}}} \right) \times 100\%
\label{eq:adp}
    \end{aligned}
\end{equation}

\noindent where \(\text{Acc}_{\text{GLUE}}\) and \(\text{Acc}_{\text{AdvGLUE}}\) is accuracy on GLUE and AdvGLUE tasks. 

We also add the metric, attack success rate (ASR),  introduced in AdvGLUE, showing the percentage of accurate predictions turned false under attacks \cite{wang2022adversarialgluemultitaskbenchmark}. 

\begin{equation}
\text{ASR} = \sum_{(x,y) \in D} \frac{\mathbf{1}[f(\mathcal{A}(x)) \neq y]}{\mathbf{1}[f(x) = y]},
\label{eq:attack_success_rate}
\end{equation}

\noindent where \(f\) is the model function, \(\mathcal{A}(x)\) is the adversarial example generated from benign sample \(x\), \(D = \{(x^{(i)}, y^{(i)})\}_{i=1}^{N}\) is the benign dataset consisting of \(N\) pairs of samples \(x\) and ground truth labels \(y\), \(\mathbf{1}[\cdot]\) be the indicator function

\section{Results and Discussion}

\renewcommand{\arraystretch}{1.2} 
\begin{table}[ht]
\centering
\caption{Finetuning GPU Memory Usage (\%)}
\label{table:efficiency}
\begin{tabular}{lcccccc}
\hline
\textbf{Model} & \textbf{SST-2} & \textbf{QQP}  & \textbf{QNLI} & \textbf{MNLI}  & \textbf{AVG}  \\
\hline
MatMul-Free LM & \textbf{29.2} & \textbf{38.2} & \textbf{62.2} & \textbf{48.1} & \textbf{43.2} \\

GLA Transformer        & 35.3 & 39.5 & 68.5 & 50.2 & 47.8 \\

Transformer++ & 50.1 & 48.8 & 74.6 & 59.4 & 57.8 \\
\hline
\end{tabular}
\vspace{0.3cm} 
\end{table}

\renewcommand{\arraystretch}{1.2} 
\begin{table}[ht]
\centering
\caption{Model Accuracy on GLUE Tasks (\%)}
\label{table:performance}
\begin{tabular}{lcccccc}
\hline
\textbf{Model} & \textbf{SST-2}  & \textbf{QQP} & \textbf{QNLI} & \textbf{MNLI}  & \textbf{AVG} \\
\hline
MatMul-Free LM & 92.9 & 83.4 & 81.7 & 75.4 & 83.5 \\

GLA Transformer        & 92.7 & 83.6 & 81.4 & 76.2 & 83.6 \\

Transformer++ & \textbf{93.2} & \textbf{84.9} & \textbf{83.3} & \textbf{78.8} & \textbf{85.8} \\
\hline
\end{tabular}
\vspace{0.3cm} 
\end{table}

\begin{table*}[ht]
    \centering
    \caption{Model Accuracy and Robustness on AdvGLUE Tasks (\%)}
    \label{table:robustness}
    \begin{tabular}{cccccc|ccccc}
        \hline
        \multirow{2}{*}{\textbf{Model}}  & \multicolumn{5}{c|}{\textbf{Acc\textsubscript{AdvGLUE}}$\uparrow$}  & \multicolumn{5}{c}{\textbf{ADP}$\downarrow$} \\ \cline{2-11}
          & SST-2 & QQP & QNLI & MNLI & AVG & SST-2 & QQP & QNLI & MNLI  & AVG \\ \hline
        MatMul-Free LM & 62.7 & 79.8 & 76.3 & 72.1 & 72.9 & 32.5 & 4.3 & 6.6 & 4.4 & 12.7 \\ 
        GLA Transformer & \textbf{65.2} & \textbf{81.5} & \textbf{76.9} & \textbf{72.7} & \textbf{74.2} & \textbf{27.5} & \textbf{0.4} & \textbf{5.6} & \textbf{3.1} & \textbf{11.2} \\ 
        Transformer++  & 61.5 & 80.2 & 75.8 & 71.2 & 72.3 & 34.0 & 5.5 & 9.0 & 9.6 & 15.7 \\ \hline
    \end{tabular}
\end{table*}

\begin{table*}[ht]
    \centering
    \caption{Model Robustness Across Different Attack Levels (\%)}
    \label{table:attack_level}
    \begin{tabular}{ccccc}
        \hline \multirow{2}{*}{\textbf{Model}}
        & \textbf{Human-level} & \textbf{Word-level} & \textbf{Sentence-level} & \textbf{AVG} \\ & Acc$\uparrow$ & ASR$\downarrow$ & ASR$\downarrow$ & ASR$\downarrow$ \\ \hline
        MatMul-Free LM & 17.8 & \textbf{5.5} & 16.5 & 9.0 \\ 
        GLA Transformer & \textbf{32.2} & 5.9 & \textbf{14.7} & \textbf{8.4} \\ 
        Transformer++ & 17.1 & 6.5 & 16.3 & 9.3 \\ \hline
    \end{tabular}
\end{table*}

\subsection{Efficiency}
Table \ref{table:efficiency} presents the memory usage of the three models on an A100 GPU. MatMul-Free LM exhibits the lowest average GPU memory usage of 43.23\%, followed by GLA at 47.75\% and Transformer++ at 57.81\%. These results align with the efficiency innovations of MatMul-Free LM, which replaces traditional matrix multiplication operations with ternary weights, and the GLA Transformer, which employs a linear complexity attention mechanism. Transformer++ consumes the most memory among the three models due to its reliance on the original Transformer architecture with additional layers and more complex attention mechanisms.

\subsection{Performance}
The performance of each model across four tasks from the GLUE benchmark is shown in Table \ref{table:performance}. Transformer++ achieves the highest overall accuracy at 85.8\% and leads in each task, underscoring its superior performance. MatMul-Free LM and GLA Transformer follow closely with an average accuracy of 83.5\% and 83.6\% respectively. Despite being optimized for efficiency, both MatMul-Free LM and GLA Transformer show only a slight decrease in performance compared to Transformer++. In addition, MatMul-Free LM and GLA Transformer are similarly effective, with GLA Transformer having a marginally higher average accuracy.

\subsection{Adversarial Robustness}
The results of the three models across AdvGLUE tasks are summarized in Table \ref{table:robustness}. We also calculate the average accuracy and ADP over all tasks for each model. The performances of the three models on GLUE and AdvGLUE exhibit distinct patterns. While Transformer++ achieves the highest overall accuracy on GLUE tasks, it has the lowest accuracy of 72.3\% on AdvGLUE and the highest Accuracy Degradation Percentage (ADP) of 15.7\%. MatMul-Free LM has comparable average accuracy on AdvGLUE but a lower average ADP compared to Transformer++, indicating potentially better adversarial robustness than Transformer++. GLA Transformer achieves the highest accuracy and the lowest ADP across all tasks, demonstrating superior robustness among the three. We also observe that ADP is much higher on SST-2 than other tasks for all three models, suggesting that the chosen models might be less robust in handling adversarial examples in sentiment analysis, which relies on subtle cues, than in other tasks, which focus more on semantic similarity and logical entailment.

Table \ref{table:attack_level} provides statistics on model resilience to different attack levels. The overall Attack Success Rate (ASR) exhibits a similar trend as the Average Decrease in Performance (ADP), with GLA Transformer having the lowest average ASR and MatMul-Free LM achieving a slightly lower ASR than Transformer++. We measure the robustness to word-level and sentence-level attacks by ASR. However, human-level perturbed cases are new samples added to AdvGLUE, which do not have corresponding samples from GLUE. Therefore, we use accuracy on the human-perturbed samples in AdvGLUE to measure robustness to human-level attacks. We observe that MatMul-Free LM tends to be the most resilient to word-level attacks among the three. One potential explanation could be that the removal of traditional matrix multiplications in attention layers obscures the gradient landscape that word-level perturbations typically exploit. GLA Transformer shows comparable robustness to word-level attacks to MatMul-Free LM and superior robustness to sentence-level and human-level attacks among the three models. This result could be explained by the linear attention and gating mechanisms of GLA Transformer, which help the model handle long sequences effectively, filter out noisy inputs, and maintain integrity in the overall context. By focusing attention more narrowly on critical parts of the input, GLA Transformer tends to resist broader and sophisticated disruptions, which aim to introduce contextually irrelevant and distracting information. Transformer++ is comparably robust to sentence-level and human-level attacks but less resilient to word-level perturbations compared to MatMul-Free LM. The complex architecture of Transformer++ helps the model excel in standard tasks by learning the pattern in data well but also could make it sensitive to small perturbations. Explanations for the varying robustness to different attack levels observed in our experimental results are based on the analysis of the model architectures. Further experiments need to be conducted to provide more detailed and reliable insights.

\subsection{Trade-off Analysis}
Our experiment results show that Transformer++ achieves high performance and decent overall adversarial robustness but comes at the cost of increased computational complexity and vulnerability to word-level attacks. MatMul-Free LM offers enhanced efficiency and robustness to word-level attacks but may struggle with maintaining detailed contextual relationships, impacting the performance on more complex tasks. GLA Transformer strikes an effective balance, providing both computational efficiency and robustness across various types of adversarial attacks while compromising little on performance. These findings demonstrate the potential of attention-efficient LLMs to provide notable robustness while improving efficiency compared to high-performance ones, making them particularly suitable for applications where resource constraints and robustness are critical. However, in scenarios where performance is crucial, models like Transformer++ may be more appropriate despite their higher computational demands. Selecting the optimal model requires careful studies and consideration of the trade-offs between efficiency, performance, and robustness to align with the context-specific needs.

\section{Conclusion}
In this study, we present a framework and conduct studies to investigate the trade-offs between efficiency, performance, and adversarial robustness of LLMs with varying mechanisms and complexities. Our analysis reveals the potential of attention-efficient models in achieving comparable performance and superior robustness compared to high-performance models with more complex architectures. These findings emphasize the importance of considering specific application requirements when selecting optimal models of deployment. 

While our framework and experiment provide valuable insights, there are several limitations. First, the efficiency evaluation is limited to fine-tuning and inference times on specific hardware, which may not fully represent real-world environments with varied infrastructure and operational costs. Second, our adversarial robustness testing is confined to limited attack methods, leaving potential vulnerabilities from newer attack strategies unaddressed. Third, we test models of specific sizes due to resource limitations, which restricts the generalization of our results across different model scales. Additionally, the framework does not account for real-world challenges such as noisy or incomplete data, system latency, or integration complexities that could affect model performance and robustness. Future research should focus on broadening the adversarial attack spectrum, applying the framework to a wider range of model sizes and architectures, and incorporating real-world operational constraints to make the framework more applicable to diverse deployment environments.

\bibliographystyle{unsrt}


\begin{thebibliography}{30}

\bibitem{vaswani2017attention}
Ashish Vaswani, Noam Shazeer, Niki Parmar, Jakob Uszkoreit, Llion Jones, Aidan~N. Gomez, \L{}ukasz Kaiser, and Illia Polosukhin.
\newblock Attention is all you need.
\newblock In {\em Proceedings of the 31st International Conference on Neural Information Processing Systems}, NIPS'17, page 6000–6010, Red Hook, NY, USA, 2017. Curran Associates Inc.

\bibitem{devlin2019bert}
Jacob Devlin, Ming-Wei Chang, Kenton Lee, and Kristina Toutanova.
\newblock Bert: Pre-training of deep bidirectional transformers for language understanding.
\newblock {\em arXiv preprint arXiv:1810.04805}, 2019.

\bibitem{brown2020language}
Tom~B. Brown, Benjamin Mann, Nick Ryder, Melanie Subbiah, Jared Kaplan, Prafulla Dhariwal, Arvind Neelakantan, Pranav Shyam, Girish Sastry, Amanda Askell, Sandhini Agarwal, Ariel Herbert-Voss, Gretchen Krueger, Tom Henighan, Rewon Child, Aditya Ramesh, Daniel~M. Ziegler, Jeffrey Wu, Clemens Winter, Christopher Hesse, Mark Chen, Eric Sigler, Mateusz Litwin, Scott Gray, Benjamin Chess, Jack Clark, Christopher Berner, Sam McCandlish, Alec Radford, Ilya Sutskever, and Dario Amodei.
\newblock Language models are few-shot learners.
\newblock {\em arXiv preprint arXiv:2005.14165}, 2020

\bibitem{johnson2020efficient}
Deepak Narayanan, Mohammad Shoeybi, Jared Casper, Patrick LeGresley, Mostofa Patwary, Vijay Korthikanti, Dmitri Vainbrand, Prethvi Kashinkunti, Julie Bernauer, Bryan Catanzaro, Amar Phanishayee, and Matei Zaharia.
\newblock Efficient large-scale language model training on gpu clusters using megatron-lm.
\newblock In {\em Proceedings of the International Conference for High Performance Computing, Networking, Storage and Analysis}, SC '21, New York, NY, USA, 2021. Association for Computing Machinery.

\bibitem{rosset2020turing}
{Microsoft}.
\newblock Turing-nlg: A 17-billion parameter language model.
\newblock {\em Microsoft Research Blog}, 2:13, 2020.

\bibitem{huang2021gshard}
Dmitry Lepikhin, HyoukJoong Lee, Yuanzhong Xu, Dehao Chen, Orhan Firat, Yanping Huang, Maxim Krikun, Noam Shazeer, and Zhifeng Chen.
\newblock Gshard: Scaling giant models with conditional computation and automatic sharding.
\newblock {\em arXiv preprint arXiv:2006.16668}, 2020.

\bibitem{zhang2020adversarial}
Shilin Qiu, Qihe Liu, Shijie Zhou, and Wen Huang.
\newblock Adversarial attack and defense technologies in natural language processing: A survey.
\newblock {\em Neurocomput.}, 492(C):278–307, 2022.

\bibitem{wan2024efficientlargelanguagemodels}
Zhongwei Wan, Xin Wang, Che Liu, Samiul Alam, Yu~Zheng, Jiachen Liu, Zhongnan Qu, Shen Yan, Yi~Zhu, Quanlu Zhang, Mosharaf Chowdhury, and Mi~Zhang.
\newblock Efficient large language models: A survey.
\newblock {\em arXiv preprint arXiv:2312.038632024}, 2024.

\bibitem{sun2023retentivenetworksuccessortransformer}
Yutao Sun, Li~Dong, Shaohan Huang, Shuming Ma, Yuqing Xia, Jilong Xue, Jianyong Wang, and Furu Wei.
\newblock Retentive network: A successor to transformer for large language models.
\newblock {\em arXiv preprint arXiv:2307.08621}, 2023.

\bibitem{gu2024mambalineartimesequencemodeling}
Albert Gu and Tri Dao.
\newblock Mamba: Linear-time sequence modeling with selective state spaces.
\newblock {\em arXiv preprint arXiv:2312.00752}, 2024.

\bibitem{yang2024gatedlinearattentiontransformers}
Songlin Yang, Bailin Wang, Yikang Shen, Rameswar Panda, and Yoon Kim.
\newblock Gated linear attention transformers with hardware-efficient training.
\newblock {\em arXiv preprint arXiv:2312.06635}, 2024.

\bibitem{zhu2024scalablematmulfreelanguagemodeling}
Rui-Jie Zhu, Yu~Zhang, Ethan Sifferman, Tyler Sheaves, Yiqiao Wang, Dustin Richmond, Peng Zhou, and Jason~K. Eshraghian.
\newblock Scalable matmul-free language modeling. \newblock {\em arXiv preprint arXiv:2406.02528}, 2024.

\bibitem{katharopoulos2020transformers}
Angelos Katharopoulos, Apoorv Vyas, Nikolaos Pappas, and François Fleuret.
\newblock Transformers are rnns: Fast autoregressive transformers with linear attention.
\newblock {\em arXiv preprint arXiv:2006.16236}, 2020.


\bibitem{choromanski2020rethinking}
Krzysztof Choromanski, Valerii Likhosherstov, David Dohan, Xingyou Song, Andreea Gane, Tamas Sarlos, Peter Hawkins, Jared Davis, Afroz Mohiuddin, Lukasz Kaiser, David Belanger, Lucy Colwell, and Adrian Weller.
\newblock Rethinking attention with performers.
\newblock {\em arXiv preprint arXiv:2009.14794}, 2022.

\bibitem{schwinn2023adversarial}
Leo Schwinn, David Dobre, Stephan G¨unnemann, and Gauthier Gidel.
\newblock Adversarial attacks and defenses in large language models: Old and new
threats.
\newblock In {\em Proceedings on}, pages 103–117, PMLR, 2023.

\bibitem{kumar2023certifying}
Aounon Kumar, Chirag Agarwal, Suraj Srinivas, Aaron Jiaxun Li, Soheil Feizi, and Himabindu Lakkaraju. \newblock Certifying llm safety against
adversarial prompting. 
\newblock {\em arXiv preprint arXiv:2309.02705}, 2023.

\bibitem{xhonneux2024efficient}
Sophie Xhonneux, Alessandro Sordoni, Stephan G¨unnemann, Gauthier Gidel, and Leo Schwinn. 
\newblock Efficient adversarial training in llms with
continuous attacks. 
\newblock {\em arXiv preprint arXiv:2405.15589}, 2024.

\bibitem{thapak2020transformer}
Prakhar Thapak and Prodip Hore.
\newblock Transformer++.
\newblock {\em arXiv preprint arXiv:2003.04974}, 2020.

\bibitem{wang-etal-2018-glue}
Alex Wang, Amanpreet Singh, Julian Michael, Felix Hill, Omer Levy, and Samuel Bowman.
\newblock {GLUE}: A multi-task benchmark and analysis platform for natural language understanding.
\newblock In Tal Linzen, Grzegorz Chrupa{\l}a, and Afra Alishahi, editors, {\em Proceedings of the 2018 {EMNLP} Workshop {B}lackbox{NLP}: Analyzing and Interpreting Neural Networks for {NLP}}, pages 353--355, Brussels, Belgium, November 2018. Association for Computational Linguistics.

\bibitem{wang2022adversarialgluemultitaskbenchmark}
Boxin Wang, Chejian Xu, Shuohang Wang, Zhe Gan, Yu~Cheng, Jianfeng Gao, Ahmed~Hassan Awadallah, and Bo~Li.
\newblock Adversarial glue: A multi-task benchmark for robustness evaluation of language models.
\newblock {\em arXiv preprint arXiv:2111.02840}, 2022.

\bibitem{deng2024composerx}
Qixin Deng, Qikai Yang, Ruibin Yuan, Yipeng Huang, Yi Wang, Xubo Liu, Zeyue Tian, Jiahao Pan, Ge Zhang, Hanfeng Lin, et al.,
\newblock ComposerX: Multi-Agent Symbolic Music Composition with LLMs,
\newblock {\em arXiv preprint arXiv:2404.18081}, 2024.

\bibitem{yang2023text}
Yumeng Yang, Soumya Jayaraj, Ethan Ludmir, and Kirk Roberts.
\newblock Text classification of cancer clinical trial eligibility criteria.
\newblock In {\em AMIA Annual Symposium Proceedings}, volume 2023, page 1304. American Medical Informatics Association, 2023.

\bibitem{liu2023chatgptpoweredconversationaldrugediting}
Shengchao Liu, Jiongxiao Wang, Yijin Yang, Chengpeng Wang, Ling Liu, Hongyu Guo, and Chaowei Xiao.
\newblock Conversational drug editing using retrieval and domain feedback.
\newblock In {\em The Twelfth International Conference on Learning Representations}, 2024.

\bibitem{ding2024largelanguagemultimodalmodels}
Jun-En Ding, Nguyen Minh Thao Phan, Wen-Chih Peng, Jian-Zhe Wang, Chun-Cheng Chug, Min-Chen Hsieh, Yun-Chien Tseng, Ling Chen, Dongsheng Luo, Chenwei Wu, Chi-Te Wang, Chih-Ho Hsu, Pei fu~Chen, Feng Liu, and Fang-Ming Hung.
\newblock Large Language Multimodal Models for New-Onset Type 2 Diabetes Prediction using Five-Year Cohort Electronic Health Records.
\newblock {\em Research Square preprint https://doi.org/10.21203/rs.3.rs-4414387/v1}, 2024. 

\bibitem{zhang2024llamafuzzlargelanguagemodel}
Hongxiang Zhang, Yuyang Rong, Yifeng He, and Hao Chen.
\newblock Llamafuzz: Large language model enhanced greybox fuzzing.
\newblock {\em arXiv preprint arXiv:2406.07714}, 2024.

\bibitem{rasheed2024aipoweredcodereviewllms}
Zeeshan Rasheed, Malik~Abdul Sami, Muhammad Waseem, Kai-Kristian Kemell, Xiaofeng Wang, Anh Nguyen, Kari Systä, and Pekka Abrahamsson.
\newblock Ai-powered code review with llms: Early results.
\newblock {\em arXiv preprint arXiv:2404.18496}, 2024.

\bibitem{wallace2019trickcanhumanintheloopgeneration}
Eric Wallace, Pedro Rodriguez, Shi Feng, Ikuya Yamada, and Jordan Boyd-Graber.
\newblock Trick me if you can: Human-in-the-loop generation of adversarial examples for question answering.
\newblock {\em arXiv preprint arXiv:1809.02701}, 2019.

\bibitem{tan2024gluepizzaeatrocks}
Zhen Tan, Chengshuai Zhao, Raha Moraffah, Yifan Li, Song Wang, Jundong Li, Tianlong Chen, and Huan Liu.
\newblock "glue pizza and eat rocks" -- exploiting vulnerabilities in retrieval-augmented generative models.
\newblock {\em arXiv preprint arXiv:2406.19417}, 2024.

\bibitem{shafahi2019adversarialtrainingfree}
Ali Shafahi, Mahyar Najibi, Amin Ghiasi, Zheng Xu, John Dickerson, Christoph Studer, Larry~S. Davis, Gavin Taylor, and Tom Goldstein.
\newblock Adversarial training for free!.
\newblock {\em arXiv preprint arXiv:1904.12843},2019.

\bibitem{tao2023metalearning}
Yiyi Tao.
\newblock Meta learning enabled adversarial defense.
\newblock In {\em Proceedings of the 2023 IEEE International Conference on Sensors, Electronics and Computer Engineering (ICSECE)}, pages 1326--1330, 2023.

\bibitem{cao2023systematic}
Meng Cao, Mehdi Fatemi, Jackie Chi~Kit Cheung, and Samira Shabanian.
\newblock Systematic rectification of language models via dead-end analysis.
\newblock {\em arXiv preprint arXiv:2302.14003}, 2023.

\bibitem{lu2024uncertainty}
Jiaying Lu, Shifan Zhao, Wenjing Ma, Hui Shao, Xiao Hu, Yuanzhe Xi, and Carl Yang.
\newblock Uncertainty-aware pre-trained foundation models for patient risk prediction via gaussian process.
\newblock In {\em Companion Proceedings of the ACM on Web Conference 2024}, WWW '24, page 1162–1165, New York, NY, USA, 2024. Association for Computing Machinery.

\bibitem{gao2018blackboxgenerationadversarialtext}
Ji Gao, Jack Lanchantin, Mary Lou Soffa, and Yanjun Qi.
\newblock Black-box generation of adversarial text sequences to evade deep learning classifiers.
\newblock {\em arXiv preprint arXiv:1801.04354}, 2018.

\bibitem{tan2024wolfwithincovertinjection}
Zhen Tan, Chengshuai Zhao, Raha Moraffah, Yifan Li, Yu~Kong, Tianlong Chen, and Huan Liu.
\newblock The wolf within: Covert injection of malice into mllm societies via an mllm operative.
\newblock {\em arXiv preprint arXiv:2402.14859}, 2024.

\bibitem{taoSQBA}
Yiyi Tao.
\newblock Sqba: sequential query-based blackbox attack.
\newblock In {\em Proceedings of the Fifth International Conference on Artificial Intelligence and Computer Science (AICS 2023)}, volume 12803, page 128032Q. International Society for Optics and Photonics, SPIE, 2017.

\bibitem{sanh2020distilbertdistilledversionbert}
Victor Sanh, Lysandre Debut, Julien Chaumond, and Thomas Wolf.
\newblock Distilbert, a distilled version of bert: smaller, faster, cheaper and lighter.
\newblock {\em arXiv preprint arXiv:1910.01108}, 2020.

\bibitem{jiao2020tinybertdistillingbertnatural}
Xiaoqi Jiao, Yichun Yin, Lifeng Shang, Xin Jiang, Xiao Chen, Linlin Li, Fang Wang, and Qun Liu.
\newblock Tinybert: Distilling bert for natural language understanding.
\newblock {\em arXiv preprint arXiv:1909.10351}, 2020.

\bibitem{Zafrir_2019}
O.~Zafrir, G.~Boudoukh, P.~Izsak, and M.~Wasserblat.
\newblock Q8bert: Quantized 8bit bert.
\newblock In {\em 2019 Fifth Workshop on Energy Efficient Machine Learning and Cognitive Computing - NeurIPS Edition (EMC2-NIPS)}, pages 36--39, Los Alamitos, CA, USA, 2019. IEEE Computer Society.

\bibitem{cho2014propertiesneuralmachinetranslation}
Kyunghyun Cho, Bart van Merri{\"e}nboer, Dzmitry Bahdanau, and Yoshua Bengio.
\newblock On the properties of neural machine translation: Encoder{--}decoder approaches.
\newblock In Dekai Wu, Marine Carpuat, Xavier Carreras, and Eva~Maria Vecchi, editors, {\em Proceedings of {SSST}-8, Eighth Workshop on Syntax, Semantics and Structure in Statistical Translation}, pages 103--111, Doha, Qatar, October 2014. Association for Computational Linguistics.

\bibitem{xu2024textsentimentanalysisclassification}
Wei Xu, Jianlong Chen, Zhicheng Ding, and Jinyin Wang.
\newblock Text sentiment analysis and classification based on bidirectional gated recurrent units (grus) model.
\newblock {\em arXiv preprint arXiv:2404.17123}, 2024.

\bibitem{hu2024minicpmunveilingpotentialsmall}
Shengding Hu, Yuge Tu, Xu~Han, Chaoqun He, Ganqu Cui, Xiang Long, Zhi Zheng, Yewei Fang, Yuxiang Huang, Weilin Zhao, et~al.
\newblock Minicpm: Unveiling the potential of small language models with scalable training strategies.
\newblock {\em arXiv preprint arXiv:2404.06395}, 2024.

\bibitem{wang2023bitnetscaling1bittransformers}
Hongyu Wang, Shuming Ma, Li~Dong, Shaohan Huang, Huaijie Wang, Lingxiao Ma, Fan Yang, Ruiping Wang, Yi~Wu, and Furu Wei.
\newblock Bitnet: Scaling 1-bit transformers for large language models. \newblock {\em arXiv preprint arXiv:2310.11453}, 2023.

\bibitem{soboleva2023slimpajama}
Daria Soboleva, Faisal Al-Khateeb, Robert Myers, Jacob~R Steeves, Joel Hestness, and Nolan Dey.
\newblock Slimpajama: A 627b token cleaned and deduplicated version of redpajama, 2023.

\bibitem{loshchilov2019decoupledweightdecayregularization}
Ilya Loshchilov and Frank Hutter.
\newblock Decoupled weight decay regularization.
\newblock {\em arXiv preprint arXiv:1711.05101}, 2019.

\end{thebibliography}

\end{document}